# TractCloud-FOV: Deep Learning-based Robust Tractography Parcellation in Diffusion MRI with Incomplete Field of View

Yuqian Chen[1,2], Leo Zekelman[2,3], Yui Lo[1,2,4], Suheyla Cetin-Karayumak[1,2], Tengfei Xue[4], Yogesh Rathi[1,2], Nikos Makris[1,5], Fan Zhang[6], Weidong Cai[4], and Lauren J. O'Donnell[1,2,7]

[1] Harvard Medical School, Boston, USA

[2] Brigham and Women's Hospital, Boston, USA

[3] Harvard University, Boston, USA

[4] The University of Sydney, Sydney, Australia

[5] Massachusetts General Hospital, Boston, USA

[6] University of Electronic Science and Technology of China, Chengdu, China

[7] Harvard-MIT Health Sciences and Technology, Cambridge, USA

**Corresponding Authors**:) Lauren J. O'Donnell (odonnell@bwh.harvard.edu) and Fan Zhang (zhangfanmark@gmail.com)

**Acknowledgments**

This work was supported by the National Institutes of Health (NIH) grants: R01MH132610, R01MH125860, R01MH119222, R01MH116173, P41EB015902, R01NS125307, R01NS125781, R01MH112748, R01AG042512, R01DC020965, R21NS136960, and K24MH116366; the National Key R&D Program (No. 2023YFE0118600); the National Natural Science Foundation (No. 62371107).

**Conflicts of Interest**

The authors declare no conflict of interest.

**Data Availability**

The HCP and ABCD datasets used in this project can be downloaded through the ConnectomeDB (db.humanconnectome.org) and ABCD Study (https://abcdstudy.org/) websites. The harmonized dMRI data from the ABCD study are available through the NIMH Data Archive (NDA) repository (https://nda.nih.gov/edit_collection.html?id=3371). The QuantConn dataset can be downloaded from the 2023 QuantConn Challenge website (http://cmic.cs.ucl.ac.uk/cdmri23/challenge.html). The ORG tractography atlas is publicly available at http://dmri.slicer.org/atlases/, and code to apply the atlas is publicly available at https://github.com/SlicerDMRI/whitematteranalysis. All code developed for our experiments will be publicly available at https://github.com/SlicerDMRI/TractCloud-FOV.

**Ethics Statement**

The creation of the WU-Minn HCP dataset was approved by the institutional review board of Washington University in St. Louis (IRB #201204036)

# Abstract

Tractography parcellation classifies streamlines reconstructed from diffusion MRI into anatomically defined fiber tracts for clinical and research applications. However, clinical scans often have incomplete fields of view (FOV) where brain regions are partially imaged, leading to partial or truncated fiber tracts. To address this challenge, we introduce TractCloud-FOV, a deep learning framework that robustly parcellates tractography under conditions of incomplete FOV. We propose a novel training strategy, FOV-Cut Augmentation (FOV-CA), in which we synthetically cut tractograms to simulate a spectrum of real-world inferior FOV cutoff scenarios. This data augmentation approach enriches the training set with realistic truncated streamlines, enabling the model to achieve superior generalization. We evaluate the proposed TractCloud-FOV on both synthetically cut tractography and two real-life datasets with incomplete FOV. TractCloud-FOV significantly outperforms several state-of-the-art methods on all testing datasets in terms of streamline classification accuracy, generalization ability, tract anatomical depiction, and computational efficiency. Overall, TractCloud-FOV achieves efficient and consistent tractography parcellation in diffusion MRI with incomplete FOV.

**Keywords**

Field of view, tractography parcellation, deep learning, diffusion MRI, data augmentation

**Key Points**

We propose a robust deep learning framework, TractCloud-FOV, to achieve robust tractography parcellation in diffusion MRI with an incomplete field of view.

We propose a novel training strategy, in which we synthetically cut tractograms to simulate real-world inferior FOV cutoff scenarios.

TractCloud-FOV outperforms state-of-the-art methods in tractography parcellation performance and efficiency on one synthetic and two real-life testing datasets.

# 1 Introduction

Diffusion magnetic resonance imaging (dMRI) is the only imaging modality that can quantify the connectivity and tissue microstructure of white matter fiber tracts, enabling their study in health and disease [Essayed et al., 2017; Nucifora et al., 2007; Zhang et al., 2022]. To perform such a study, the segmentation or parcellation of white matter fiber tracts is a critical prerequisite task. However, parcellating these tracts typically relies on high-quality research dMRI acquisitions where fiber tracts are fully traceable. Existing tools often fail on clinical scans with incomplete fields of view (FOV), where peripheral brain regions may be partially imaged, leading to partial or truncated fiber tracts.

Inferior or ventral FOV cutoff, where crucial regions of the brainstem and cerebellum are missing from a dMRI scan, poses an important challenge. This artifact, illustrated in Figure 1, can prevent identification of key tracts that traverse the cerebellum and brainstem, such as the corticospinal tract, cortico-ponto-cerebellar pathway, and inferior cerebellar peduncle. Inferior FOV cutoff is common in clinically acquired dMRI data and affects large-scale research datasets, such as the Adolescent Brain Cognitive Development Study, where automated post-processing [Hagler et al., 2019] indicates nearly half of scans have some degree of ventral FOV cutoff.

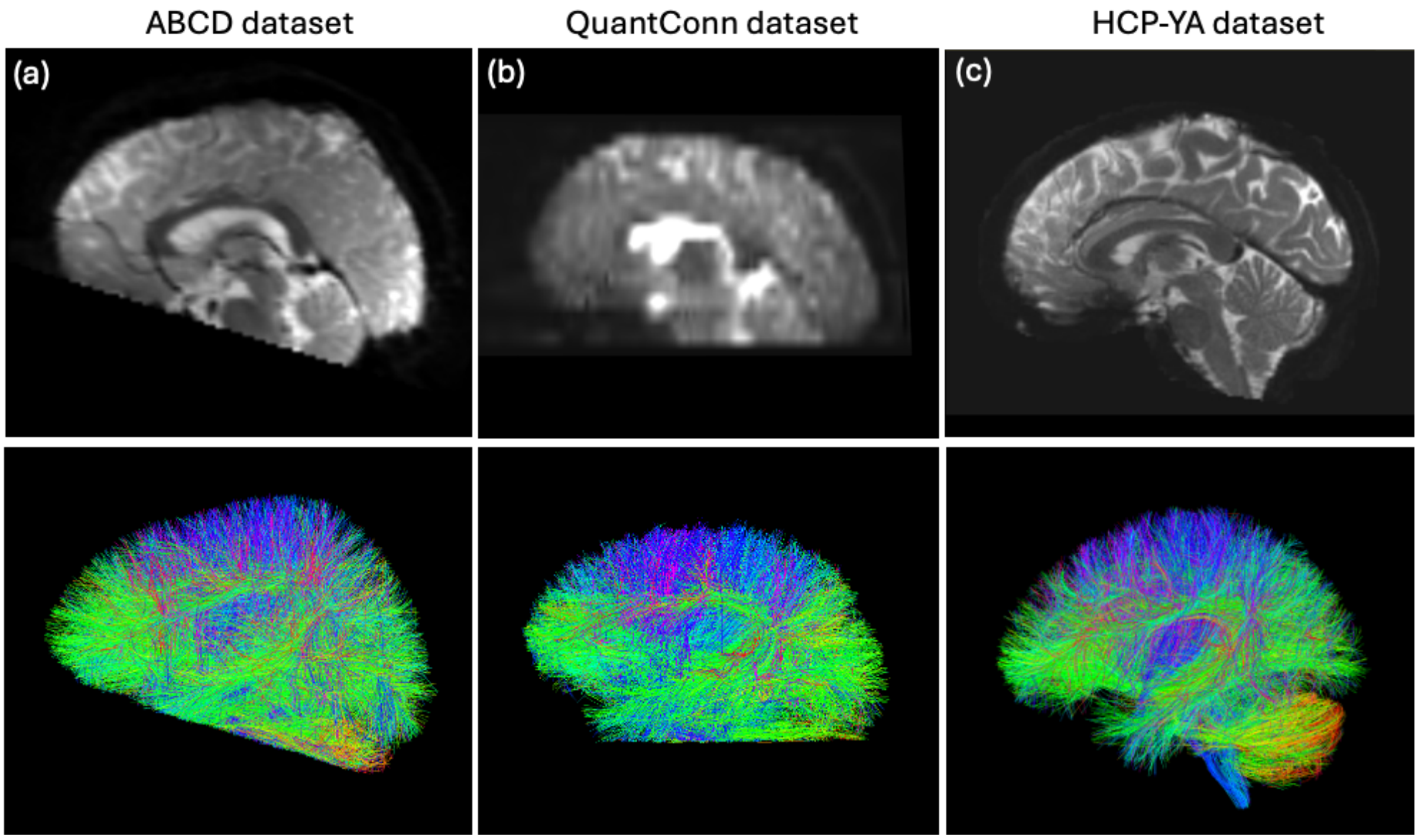

Figure 1. Incomplete FOV DWI scans and tractograms from example subjects in the ABCD-FOV dataset [Cetin-Karayumak et al., 2024] (a) and QuantConn dataset [Newlin et al., 2024] (b), compared to a complete DWI scan and tractogram of an example subject from the HCP-YA dataset [Van Essen et al., 2013] (c).

In the analysis of structural MRI scans (such as T1-weighted or T2-weighted), imaging artifacts such as incomplete FOV acquisitions can be addressed by robust deep learning segmentation methods. For example, SynthSeg+ enables robust clinical MRI segmentation and is trained using a synthetic data strategy that simulates a wide range of possible clinical image acquisitions and artifacts [Billot et al., 2023]. However, in dMRI, robust tractography parcellation methods remain underdeveloped. Recent studies have addressed the dMRI FOV cutoff issue by using deep learning to impute missing diffusion signal data for individual subjects [Gao et al., 2024b; Li et al., 2024]. This imputed data fills in image gaps to allow more complete tractography, which can then be parcellated using standard methods.

However, imputed dMRI data lacks subject specificity, potentially compromising metrics like fractional anisotropy, which depend on accurate diffusion properties to assess brain health.

In this paper, we introduce TractCloud-FOV, a deep learning framework that robustly parcellates tractography under conditions of incomplete FOV. TractCloud-FOV aims to fully and accurately identify all subject-specific tracts within each subject's acquired field of view, rather than imputing synthetic dMRI signals. To achieve this, we introduce a novel data augmentation technique during the training stage, termed FOV-Cut Augmentation (FOV-CA). FOV-CA simulates varying degrees of inferior FOV cutoff by synthetically cutting tractograms. This augmentation strategy enhances the model's ability to generalize across diverse incomplete FOV scenarios by providing realistic training examples of truncated streamlines. FOV-CA enables robust tractography parcellation in real-world scenarios without the need for image imputation. We implement FOV-CA as an extension to our recently proposed geometric deep learning framework, TractCloud, which employs a local-global streamline representation for enhanced classification [Xue et al., 2023a]. The extended method, TractCloud-FOV, is a deep learning framework designed to robustly parcellate tractography derived from dMRI data with incomplete FOV. The key contributions of this work are:

1. Robust Parcellation Framework: A deep learning approach for effective parcellation of tractograms with inferior FOV cutoff.
2. Synthetic Data Augmentation: A training strategy that employs synthetically cut tractograms to simulate various FOV cutoff scenarios.
3. Real-World Validation: A comprehensive evaluation of the framework using both simulated cutoff data and real-world datasets with partial FOV acquisitions.

Our results demonstrate that TractCloud-FOV successfully addresses challenges posed by incomplete FOV in dMRI, achieving robust and accurate tractography parcellation.

# 2 Methods

## 2.1 dMRI Tractography Datasets

To train and evaluate the effectiveness of our TractCloud-FOV method, we investigate dMRI data from three datasets that were independently acquired from distinct populations using different imaging protocols and scanners.

### 2.1.1 HCP-YA Dataset (Train/Validation/Test)

We use a tractography dataset of 1 million labeled streamlines from the O'Donnell Research Group (ORG) Atlas [Zhang et al., 2018], which is available online[1]. The streamline labels include 74 tract classes: 73 anatomically meaningful tracts from the whole brain and one tract category of "other streamlines" including anatomically implausible outlier streamlines. These labels were anatomically annotated by an expert neuroanatomist who viewed the 1 million streamlines organized into population-based fiber clusters [Zhang et al., 2018]. These streamlines were initially created by applying a two-tensor unscented Kalman filter tractography method (UKFt) [Reddy and Rathi, 2016] to generate tractograms from the dMRI data of 100 healthy young adults from the minimally preprocessed Human Connectome Project Young Adult (HCP-YA) dataset [Van Essen et al., 2013]. HCP-YA dMRI scans used multiband EPI (slice acceleration factor=3, 1.25 mm isotropic resolution, 111 oblique axial image slices spanning approximately 139 mm in the superior-inferior direction and covering the cerebrum, cerebellum, and brainstem) with 18 b=0 volumes and three b-values (b= 1000, 2000, 3000 s/mm²) with 90 directions each, acquired with TR/TE=5520/89.5 ms, followed by minimal preprocessing including eddy current, motion, and distortion correction [Glasser et al., 2013]. The b=3000 shell and b=0 volumes were used for UKFt, as the b=3000 shell is optimal for resolving crossing fibers [Chen et al., 2025; Descoteaux et al., 2007; Ning et al., 2015; Zekelman et al., 2022]. As a deterministic tractography algorithm, UKFt employs a multi-fiber model to handle crossing fibers and stabilizes tracking by incorporating prior information at each step [Reddy and Rathi, 2016], as implemented in the ukftractography package[2]. The use of a multi-fiber model for reconstruction of crossing fibers is important for the depiction of tracts affected by inferior FOV cutoff. Such tracts include the corticospinal tract, whose lateral connections are crossed by the superior longitudinal fasciculus, and the cortico-ponto-cerebellar tract, which decussates at the level of the pons. UKFt consistently reconstructs white matter anatomy across the lifespan and across different acquisitions [Zhang et al., 2018] and has high sensitivity for reconstructing lateral parts of the corticospinal tract [He et al., 2023]. The 1 million labeled streamlines include 10,000 streamlines from each of the 100 HCP-YA subjects. We divide this labeled HCP-YA tractography data into train/validation/test sets with the distribution of 70%/10%/20% by subjects (such that all streamlines from an individual subject are placed into only one set, either train or validation or test, to prevent data leakage of subject-specific information across sets [Xue et al., 2023a]).

[1] https://github.com/SlicerDMRI/ORG-Atlases

### 2.1.2 ABCD-FOV Dataset (Test Data)

We investigate tractography from the publicly available harmonized Adolescent Brain Cognitive Development (ABCD) dMRI dataset [Cetin-Karayumak et al., 2024]. The ABCD dMRI data was first harmonized to remove scanner-specific effects across 21 acquisition sites, followed by UKFt, and the dMRI and tractography were released on the National Institute of Mental Health Data Archive [Cetin-Karayumak et al., 2024; Chen et al., 2023b; Xue et al., 2024]. Previously, the dMRI data was originally acquired (b = 500, 1000, 2000 and 3000 s/mm$^2$, resolution = 1.7 × 1.7 × 1.7 mm$^3$), minimally preprocessed (eddy and motion correction, b0 inhomogeneity correction, gradient unwarp, and resampling to isotropic resolution), and quality controlled by the ABCD study Data Analysis, Informatics, and Resource Center group [Casey et al., 2018]. The ABCD dMRI dataset uses an axial acquisition with 81 image slices covering approximately 138 mm in the superior-inferior direction. This is usually sufficient to cover the cerebrum, cerebellum, and brainstem, but incomplete FOV can result due to factors such as subject positioning and operator-dependent acquisition.

For inclusion in our study, we create an ABCD-FOV dataset by selecting harmonized ABCD tractography data with incomplete FOV from 80 children (ages 9.9 ± 0.6 years, 23 females). This dataset represents a challenging scenario for tractography parcellation, where a research dMRI acquisition is affected by varying levels of incomplete FOV. Participants are selected to encompass several levels of inferior incomplete FOV artifact (mild, moderate, and severe), according to automated post-processing quality control metrics released by the ABCD study (table: mri_y_qc_auto_post)[3]. Specifically, FOV cutoff was quantified by ABCD using two measures: dorsal (superior) cutoff (*apqc_dmri_fov_cutoff_dorsal*) and ventral (inferior) cutoff (*apqc_dmri_fov_cutoff_ventral*), calculated as the percent intersection of the brain mask with frame borders. We exclude participants with dorsal (superior) cutoff scores above zero since this study focuses on addressing the challenge of ventral (inferior) cutoff. For ventral (inferior) cutoff, which ranges from 0 (no cutoff) to 53 (high cutoff), we categorize participants into three groups: mild (15 ≤ cutoff ≤ 24), moderate (25 ≤cutoff ≤ 34), and severe (cutoff ≥ 35). We randomly select participants to include 40 with severe cutoff, 20 with moderate cutoff, and 20 with mild cutoff, resulting in a total of 80 participants with varying levels of incomplete FOV artifact.

[2] https://github.com/pnlbwh/ukftractography.

[3] https://wiki.abcdstudy.org/release-notes/imaging/quality-control.html#automated-post-processing-qc

Figure 1(a) shows a DWI scan and corresponding tractogram from an example ABCD-FOV subject, while Figure 2 illustrates example tractograms for participants with mild, moderate, and severe cutoff.

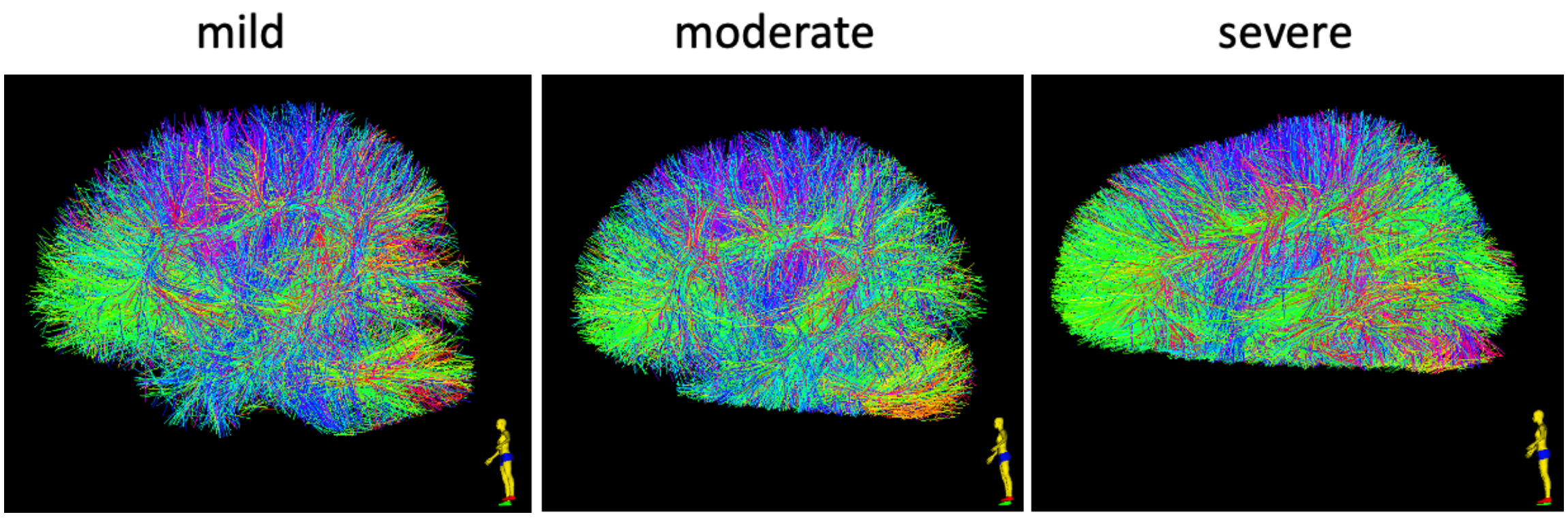

Figure 2. Example whole brain tractograms for participants with mild, moderate, and severe incomplete FOV artifacts from the ABCD dataset.

### 2.1.3 QuantConn Dataset (Test Data)

We also investigate the QuantConn challenge dataset [Newlin et al., 2024], which is a subset of the Queensland Twin Imaging study [Strike et al., 2023]. This dataset presents a difficult scenario for tractography parcellation, featuring a clinical style, low-resolution dMRI acquisition with prevalent incomplete FOV artifacts. DWI images were acquired using single-shot echo-planar imaging (b = 0 and 1146 s/mm$^2$, resolution = 1.8 × 1.8 × 5 mm$^3$). Each 3D DWI volume was an axial acquisition of 21 image slices with 5 mm slice thickness and 0.5 mm gap. This acquisition covers about 115 mm in the superior-inferior direction and is designed to cover the cerebrum with generally partial coverage of the brainstem and cerebellum, resulting in incomplete FOV. The DWI data were processed using the PreQual pipeline to remove eddy current, motion, and EPI distortions [Cai et al., 2021]. We randomly select 50 subjects (ages 25.3 ± 1.8 years, 22 females) for inclusion in this study. We obtain whole brain tractography from the dMRI data using UKFt. Figure 1(b) shows a DWI scan and corresponding tractogram from an example QuantConn subject.

## 2.2 Network Architecture

In this study, we adopt the TractCloud network architecture proposed in our previous work [Xue et al., 2023a] to perform streamline classification. TractCloud, a point-based neural network, processes streamlines as point clouds and uniquely leverages a local-global streamline representation. Unlike traditional point-based networks [Qi et al., 2017; Wang et al., 2019], TractCloud learns a local-global streamline representation that leverages information from neighboring and whole-brain streamlines to improve classification performance. It also uses data augmentation with synthetic transforms during training to enable registration-free tractography parcellation.

The TractCloud architecture comprises two main parts: a local-global representation learning module and a point-based neural network for streamline classification. The input to the representation learning module is constructed by concatenating the coordinates of each streamline with its local neighbor streamlines and global whole-brain streamlines. The representation learning module begins with a shared fully connected layer with ReLu activation function, followed by a max-pooling layer. The output of this module is the learned local-global representation of an input streamline. Then the representation is input to the point-based neural network (PointNet [Qi et al., 2017] in this study) for classification. The representation learning module replaces the first layer of the point-based network, and the whole network including the two parts is trained in an end-to-end fashion.

## 2.3 FOV-Cut augmentation with synthetically cut streamlines

Our previous TractCloud training strategy utilized synthetically transformed tractograms to achieve registration-free tractography parcellation. In this study, we extend this approach to handle incomplete streamlines by introducing a novel data augmentation method that we call FOV-Cut Augmentation (FOV-CA). Instead of applying standard synthetic transforms (e.g., rotation, scaling as in [Xue et al., 2023a]), we simulate varying degrees of inferior FOV cutoff by synthetically cutting tractograms, generating augmented data for training, validation, and testing. This strategy enables robust model training and evaluation under conditions mimicking real-world incomplete FOV cutoff challenges.

For FOV-Cut Augmentation, we apply cutting planes to our dataset of 1 million labeled streamlines from 100 HCP-YA subjects. Planes are randomly generated and are constrained to 30–50 mm below the center of the brain with an angle of under 30° to the horizontal plane, simulating realistic FOV cutoff scenarios, especially affecting the brainstem and cerebellum (Figure 3). On average, this FOV-CA process cuts 10% of the streamlines in each tractogram (Figure 4). The cutting process affects 26 anatomical tracts located in or passing through inferior brain regions[4].

Using FOV-CA, the HCP-YA labeled streamline training data (70 subjects) are augmented by applying 10 different cutting planes to each whole-brain tractogram, yielding 10 synthetically cut tractograms per subject (plus the original input tractogram, for a total of 11 tractograms per training subject). This augmentation results in a total of approximately 7.6 million streamlines for training (70 subjects x 10,000 streamlines per tractogram x 11 tractograms gives 7.7 million labeled streamlines, minus a small number of streamlines that are removed entirely due to their location below the cutting plane, resulting in approximately 7.6 million training streamlines).

The HCP-YA independent validation (10 subjects) and testing (20 subjects) tractograms are also synthetically augmented using FOV-CA to simulate real-world incomplete FOV challenges. Here, only one synthetically cut tractogram is generated for each validation/testing subject, resulting in one "original complete" and one "synthetically cut" version for each subject. These two versions are used for performance evaluation. In addition, each synthetically cut tractogram includes streamlines categorized as "cut" or "unaffected" by the applied cutting plane, which are used for performance evaluation.

[4] Abbreviation lists of affected tracts: T_CPC, T_CR-F_left, T_CR-F_right, T_CR-P_left, T_CR-P_right, T_CST_left, T_CST_right, T_ICP_left, T_ICP_right, T_Intra-CBLM-I-P_left, T_Intra-CBLM-I-P_right, T_Intra-CBLM-PaT_left, T_Intra-CBLM-PaT_right, T_MCP, T_Sup-OT_left, T_Sup-OT_right, T_Sup-O_left, T_Sup-O_right, T_Sup-PO_left, T_Sup-PO_right, T_Sup-PT_left, T_Sup-PT_right, T_Sup-T_left, T_Sup-T_right, T_UF_left, T_UF_right. Refer to https://github.com/SlicerDMRI/ORG-Atlases/blob/master/Tracts-in-ORG-800FC-100HCP.md for the definitions of these tracts.

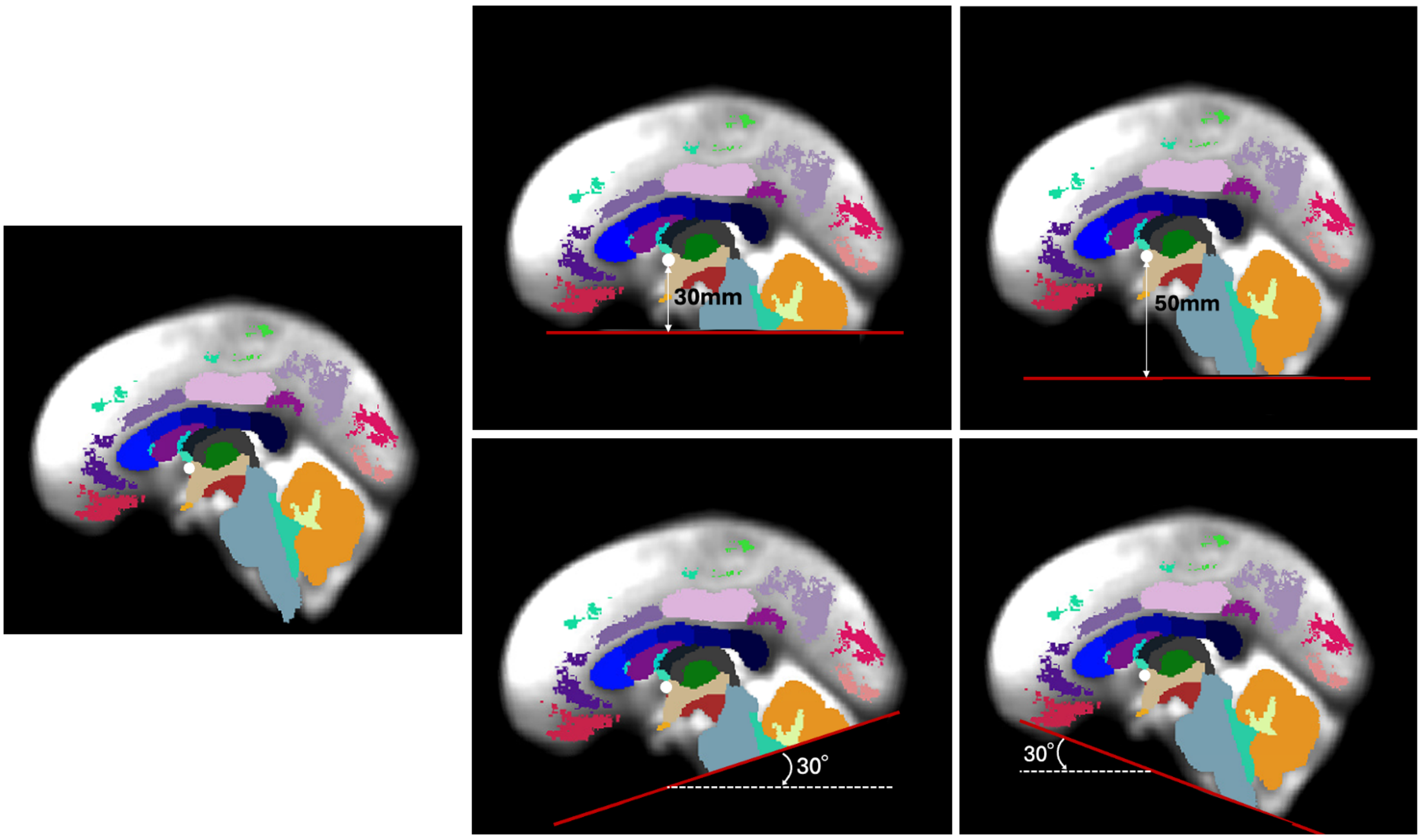

Figure 3. Illustration of locations of cutting planes used for synthetic data generation. The background is the ORG atlas b0 image with FreeSurfer [Fischl, 2012] regions overlaid in color. The brainstem (gray) and cerebellum (orange and yellow) regions are especially affected by the synthetic cutting process.

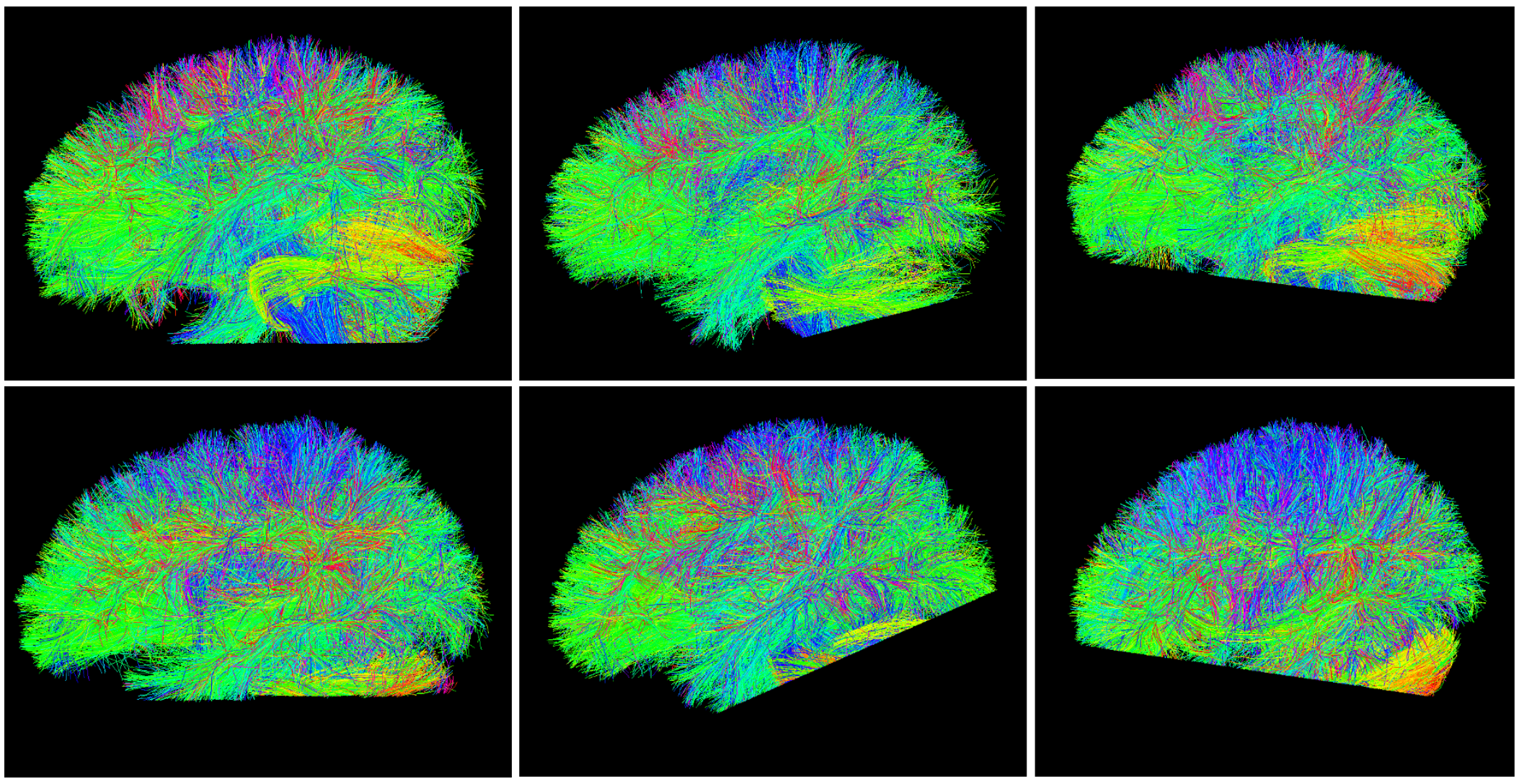

Figure 4. Examples of synthetically cut whole brain tractograms from HCP-YA subjects.

## 2.4 Implementation details

The training of our TractCloud-FOV framework took about 5.6 hours and 11 GB GPU memory on an NVIDIA RTX A100 GPU using Pytorch (v1.12.1) [Paszke et al., 2019]. Our overall network is trained for 20 epochs with a learning rate of 0.001. The batch size of training is 1024 and Adam [Kingma and Ba, 2014] is used for optimization using cross-entropy loss. For the number of local and global streamlines, we use the default setting in [Xue et al., 2023a] (20 local and 500 global streamlines).

During the synthetic augmentation of training data, the number of cutting planes applied to each training subject is set to 10. For the independently acquired testing datasets (ABCD-FOV and QuantConn) the average number of streamlines per subject in the whole brain tractography is approximately 500,000. All whole brain tractography is registered to the ORG atlas space before tractography parcellation [Chen et al., 2022; Zhang et al., 2018].

# 3 Experiments and Results

## 3.1 Performance on the Labeled HCP-YA Dataset

We train a classification model using the synthetic training data from Section 2.4 and evaluate its performance on the labeled HCP-YA testing data, including both original complete and synthetically cut tractography. The proposed method is compared against two state-of-the-art (SOTA) deep learning models: PointNet [Qi et al., 2017] and the original TractCloud [Xue et al., 2023a]. PointNet is a classical point-based neural network for processing point clouds and has been successfully applied to tractography parcellation [Chen et al., 2023a; Xue et al., 2023b]. T-Net (the spatial transformation layer) is removed from PointNet to preserve anatomically important information about the spatial position of streamlines [Chen et al., 2024]. TractCloud was proposed in our previous study to perform tractography parcellation using a local-global streamline representation [Xue et al., 2023a]. For all compared methods in this study, the tractograms are registered to the ORG atlas space [Zhang et al., 2018]. Thus the synthetic transform augmentation proposed in the original TractCloud for registration-free parcellation [Xue et al., 2023a] is not included at this stage. Our TractCloud-FOV method extends TractCloud by incorporating synthetically cut streamline augmentation (FOV-CA) during training. Default parameters from the respective studies are used for all SOTA methods. Tractography parcellation is performed via streamline classification, with performance evaluated using two metrics, accuracy and macro F1 score. The accuracy is calculated as the overall accuracy of all testing streamlines, and the macro F1 score is calculated as the mean across 73 tract classes. Table 1 provides streamline classification performance for all compared methods. The performance metrics are reported on original complete and synthetically cut tractography. Additionally, within the synthetically cut tractography, we separately evaluate classification performance on streamlines that are cut and those that are unaffected by the cutting plane.

Table 1. Performance comparison of streamline classification accuracy and F1 scores across methods.

| Experiments | original complete tractography | | synthetically cut tractography | | | | | |
|---|---|---|---|---|---|---|---|---|
| | | | all streamlines | | cut streamlines | | unaffected streamlines | |
| | acc | F1 | acc | F1 | acc | F1 | acc | F1 |
| PointNet | 91.10 | 88.37 | 89.00 | 83.99 | 76.16 | 66.85 | 90.93 | 87.97 |
| TractCloud | 91.78 | 89.70 | 89.42 | 84.95 | 75.97 | 67.43 | 91.44 | 89.03 |
| TractCloud-FOV | **94.11** | **92.57** | **92.85** | **90.24** | **85.91** | **79.18** | **93.89** | **91.99** |

As shown in Table 1, our proposed TractCloud-FOV method outperforms SOTA methods in terms of both accuracy and F1 score. For the complete tractography, the superior performance may be attributed to the higher number of training streamlines generated through synthetic augmentation. For the synthetically cut tractography, our method achieves the best results. A detailed analysis of cut and unaffected streamlines within the synthetically cut tractography reveals a notable improvement in the classification of cut streamlines by TractCloud-FOV (nearly 10% higher accuracy compared to SOTA methods) as well as an improvement in the classification of unaffected streamlines.

## 3.2 Evaluation on Unseen Test Data

To assess the robustness and generalization of our trained model, we conducted experiments on two independently acquired datasets: ABCD-FOV (80 subjects) and QuantConn (50 subjects). Both datasets consist of tractography derived from DWI scans with incomplete FOV. We compared our method against three SOTA tractography parcellation methods: RecoBundles [Garyfallidis et al., 2018], WMA [Zhang et al., 2018], and TractCloud [Xue et al., 2023a]. RecoBundles leverages bundle models as shape priors to detect and group similar streamlines into white matter tracts. For this comparison experiment, RecoBundles used the ORG atlas to provide shape priors. WMA is an atlas-based clustering method with high cross-subject correspondence, also utilizing the ORG atlas.

We evaluated performance using two established metrics for evaluation of tractography parcellation performance in the absence of ground truth. The Tract Identification Rate (TIR) measures the percentage of tracts successfully identified in a subject, with a minimum of 20 streamlines required for a tract to be considered identified [Chen et al., 2021; Chen et al., 2023a; Zhang et al., 2018]. TIR was averaged across all subjects. The Atlas-to-Tract Distance (ATD) measures the geometric similarity between identified tracts and corresponding tracts from the training atlas [Xue et al., 2023b]. For each tract, the average pointwise distance between the atlas and identified tracts was computed and averaged across subjects and tracts.

For performance evaluation, we assessed results in the whole tractogram (all tracts) as well as in the tracts expected to be most affected by inferior FOV cutoffs (affected tracts, see Section 2.3 for details). We analyzed the performance of parcellating the affected tracts to evaluate the effectiveness of our method on incomplete streamlines. TIR and ATD results for all tracts and affected tracts are shown in Tables 2–5 (the numbers in parentheses indicate standard deviations across subjects) for the ABCD-FOV and QuantConn datasets. To conduct statistical analysis on the results of evaluation metrics, a one-way repeated measures Analysis of Variance (ANOVA) was applied to each of the two evaluation metrics, followed by post-hoc pairwise comparisons using paired t-tests between our proposed method and each comparison method.

To further evaluate performance on ABCD-FOV, we analyzed subject groups based on FOV cutoff severity (mild, moderate, severe). TIR and ATD metrics for these groups are presented in Tables 2 and 4, respectively. Similarly, statistical analysis of ANOVA followed by paired t-tests was applied to each subject group.

Table 2. Tract identification rate (TIR, higher is better) of all tracts and affected tracts for ABCD-FOV data across methods.

| Method | RecoBundles | WMA | TractCloud | TractCloud-FOV |
|---|---|---|---|---|
| all tracts | 0.974 (0.049)** | 0.978 (0.032)** | 0.983 (0.025)** | **0.992 (0.017)** |
| affected tracts | 0.931 (0.091)** | 0.940 (0.089)** | 0.961 (0.061)** | **0.997 (0.012)** |
| affected tracts (mild) | 0.983 (0.051)** | 0.992 (0.015) | **0.996 (0.012)** | **0.996 (0.017)** |
| affected tracts (moderate) | 0.967 (0.052)** | 0.994 (0.014) | 0.988 (0.021)* | **0.996 (0.012)** |
| affected tracts (severe) | 0.888 (0.099)** | 0.888 (0.100)** | 0.930 (0.071)** | **0.997 (0.010)** |

*p<0.05 **p<0.0001

Table 3. Tract identification rate (TIR, higher is better) of all tracts and affected tracts for QuantConn data across methods.

| Method | RecoBundles | WMA | TractCloud | TractCloud-FOV |
|---|---|---|---|---|
| all tracts | 0.925 (0.038)** | 0.941 (0.034)** | 0.959 (0.028) * | **0.981 (0.029)** |
| affected tracts | 0.806 (0.086)** | 0.839 (0.084)** | 0.897 (0.069)** | **0.962 (0.061)** |

*p<0.05 **p<0.0001

Table 4. Atlas-to-tract distance (ATD, lower is better) of all tracts and affected tracts for ABCD-FOV data across methods.

| Method | RecoBundles | WMA | TractCloud | TractCloud-FOV |
|---|---|---|---|---|
| all tracts | 1.591 (0.535) | 1.727 (0.471)** | 1.734 (0.464)** | **1.563 (0.420)** |
| affected tracts | 2.551 (1.176)** | 2.484 (1.009)** | 2.262 (0.969)** | **1.968 (0.772)** |
| affected tracts (mild) | 1.761 (0.700)* | 1.692 (0.468)** | 1.528 (0.465)* | **1.427 (0.442)** |
| affected tracts (moderate) | 1.943 (0.799)** | 1.848 (0.649)** | 1.725 (0.586)* | **1.513 (0.431)** |
| affected tracts (severe) | 3.251 (1.084)** | 3.199 (1.020)** | 2.898 (0.869)** | **2.467 (1.397)** |

*p<0.05 **p<0.0001

Table 5. Atlas-to-tract distance (ATD, lower is better) of all tracts and affected tracts for QuantConn data across methods.

| Method | RecoBundles | WMA | TractCloud | TractCloud-FOV |
|---|---|---|---|---|
| all tracts | 2.501 (0.429)** | 2.299 (0.34)** | 2.237 (0.329)** | **2.110 (0.274)** |
| affected tracts | 4.104 (0.777)** | 3.852 (0.859)** | 3.284 (0.628)** | **3.119 (0.555)** |

*p<0.05 **p<0.0001

As shown in Tables 2–5, our TractCloud-FOV method achieves superior tractography parcellation performance across all tracts and affected tracts in both ABCD-FOV and QuantConn datasets, with significantly higher TIR and significantly lower ATD compared to SOTA methods ($p<0.05$ in all analyses, as indicated by asterisks in Tables 2-5). The high TIR demonstrates the strong generalization ability of our model, while the low ATD indicates superior consistency between identified tracts and the atlas. Notably, our method shows large improvements for FOV-affected tracts, demonstrating its effectiveness in parcellating incomplete FOV tractography.

The ABCD-FOV subgroup results indicate that for mild and moderate incomplete FOV data, TractCloud-FOV achieves the highest tract identification rate (TIR), while TractCloud and WMA also perform reasonably well, with WMA showing no significant difference from TractCloud-FOV.

However, in the presence of severe incomplete FOV artifacts, TractCloud-FOV significantly outperforms all SOTA methods in TIR (p<0.0001 in all analyses). Additionally, in terms of atlas consistency (ATD), TractCloud-FOV significantly outperforms all SOTA methods across mild, moderate, and severe FOV cutoff groups (p<0.05 in all analyses).

We provide visualizations of three example identified tracts from affected tracts across comparison methods for both datasets in Figures 5 and 6. It can be seen that the proposed TractCloud-FOV method identifies more complete tracts than comparison methods, with tracts more closely resembling the corresponding tracts in the ORG atlas.

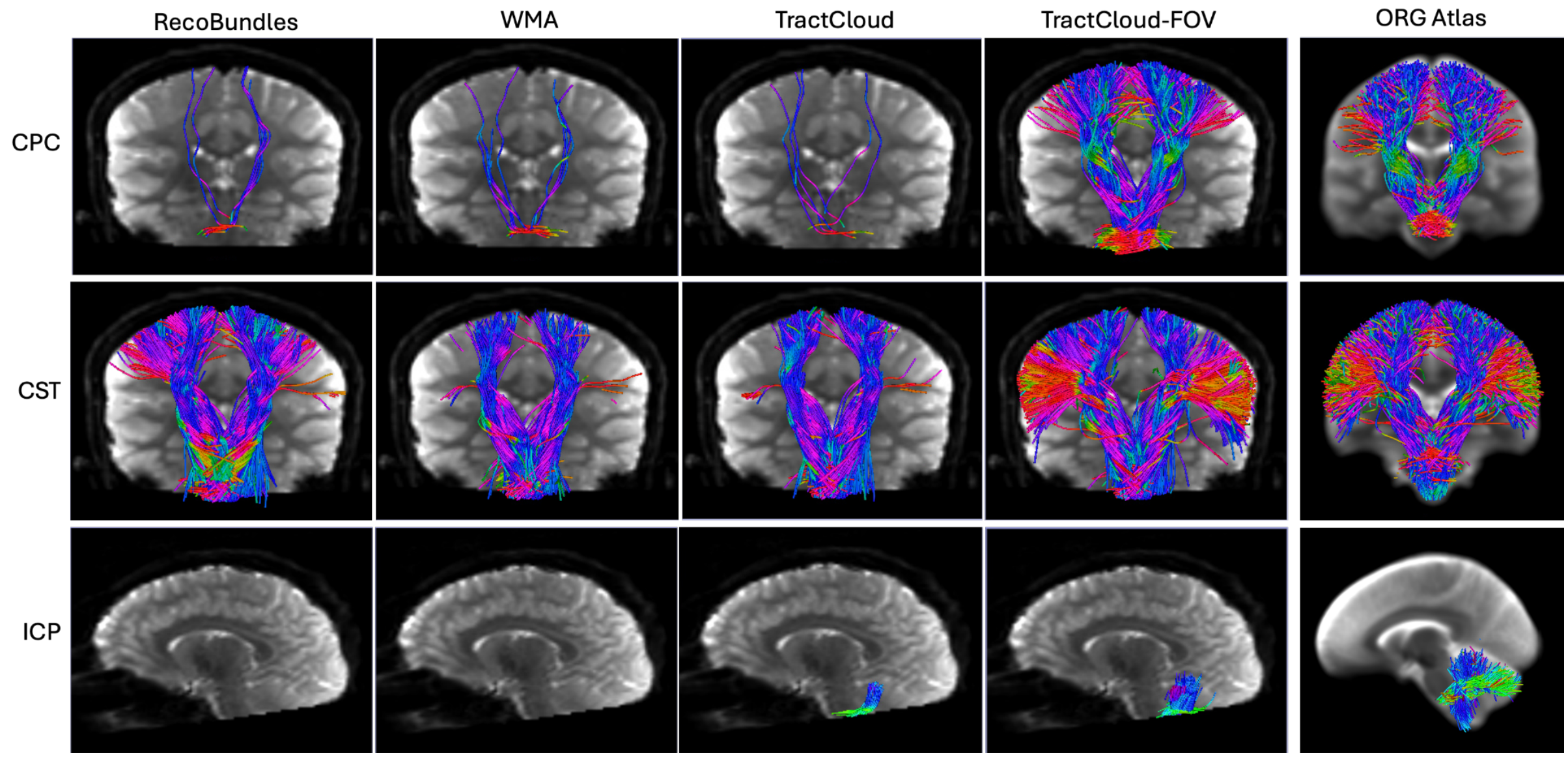

Figure 5. Visualization of three example tracts generated by four tractography parcellation methods applied to an ABCD-FOV subject and the corresponding tracts from the ORG atlas (the rightmost column). The views of the three rows are anterior, anterior and left, respectively. (CPC: cortico-ponto-cerebellar, CST: corticospinal tract, ICP: inferior cerebellar peduncle)

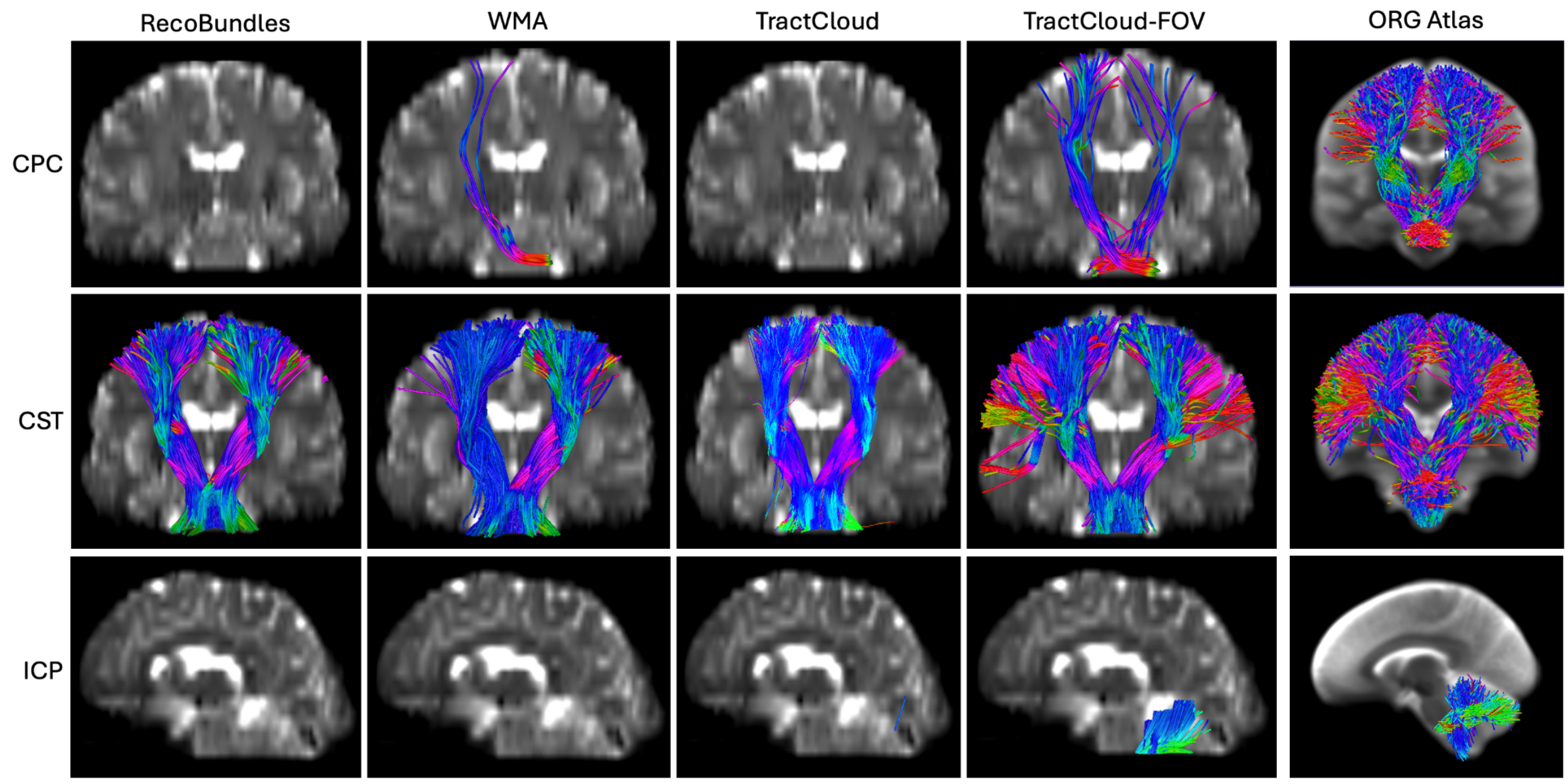

Figure 6. Visualization of three example tracts generated by four tractography parcellation methods applied to a QuantConn subject and the corresponding tracts from the ORG atlas (the rightmost

column). The views of the three rows are anterior, posterior and left, respectively. (CPC: cortico-ponto-cerebellar, CST: corticospinal tract, ICP: inferior cerebellar peduncle)

To assess efficiency, the computation time for each comparison method to perform tractography parcellation on a single QuantConn subject was recorded (Table 6). All methods were tested on a system with a 2.1 GHz Intel Xeon E5 CPU (8 DIMMs, 32 GB memory). The results indicate that TractCloud-FOV and TractCloud are much faster than WMA and slightly outperform RecoBundles in terms of speed.

Table 6. Computation time comparison

| Method | RecoBundles | WMA | TractCloud | TractCloud-FOV |
|---|---|---|---|---|
| time (s) | 182 | 2510 | 98 | 108 |

# 4 Discussion and Conclusion

In this study, we introduce TractCloud-FOV, a deep learning framework for robust tractography parcellation of dMRI data with an incomplete field of view. Our approach introduces a unique training strategy, FOV-Cut Augmentation, that employs data augmentation using synthetically cut tractograms to simulate FOV cutoffs. This training strategy enables TractCloud-FOV to address challenges posed by incomplete data. TractCloud-FOV demonstrates superior effectiveness in comparison with SOTA methods on simulated and real-world data. Real-world results show significant improvements for FOV-affected tracts and smaller but consistent gains for all tracts. Overall, these findings indicate that TractCloud-FOV not only enhances tractography parcellation for incomplete FOV tractograms but also improves performance for complete tractograms, demonstrating its robustness and broad applicability.

To our knowledge, this is the first work to investigate tractography parcellation of DWI with incomplete FOV. Previous studies have investigated the problem of dMRI with incomplete FOV [Gao et al., 2024a; Li et al., 2024]. They proposed methods to impute missing slices from existing dMRI scans before conducting downstream tractography analysis. In contrast, our method directly performs tractography parcellation on incomplete DWI data, eliminating the need for image imputation. In this way, our approach aims to fully and accurately identify all tracts present within each subject's limited FOV.

Incomplete FOV is one important aspect of clinical acquisition that affects fiber tract reconstruction. Other factors, such as acquisition resolution or b-value, choice of tractography approach, and choice of tractography parcellation method, can also impact fiber tract reconstruction [Schilling et al., 2021]. Despite these challenges, robust technology for identification of major bundles across different acquisitions has now been available for several years [Garyfallidis et al., 2018; Wasserthal et al., 2018; Zhang et al., 2018]. However, even when using multi-fiber tractography methods, crossing or decussating fibers still pose challenges to fully reconstruct several tracts studied here including the corticospinal and cortical-ponto-cerebellar tracts, as well as the superior cerebellar peduncle [He et al., 2023; Lundell and Steele, 2024]. Future work towards robust parcellation of clinical tractography may consider the impact of these many factors in the setting of incomplete FOV acquisitions.

The experimental results in Table 1 show that our proposed TractCloud-FOV not only benefits the performance of cut streamlines, but also benefits unaffected streamlines and original complete tractograms. The reason is likely that our proposed FOV-Cut Augmentation strategy greatly increased the training data (from 700 thousand to 7.6 million streamlines) by generating synthetically cut tractograms. Though streamlines unaffected by cutting planes may be repeated within the multiple cut training tractograms for each subject, these repeated streamlines actually result in distinct training samples for the streamline classification model. This is because when using the TractCloud model, different local-global representations are learned from the corresponding whole-brain tractogram for each streamline. Because each tractogram is synthetically cut in a different way, this affects the local-global representation and allows even repeated streamlines to serve as distinct training samples for the training process.

Finally, we note some limitations of this study and directions for future research. First, this study focused on inferior or ventral FOV cutoffs, which are common in clinical and research datasets. However, superior FOV cutoffs can also occur and should be addressed in future work by incorporating additional training data augmentation. This future work can leverage testing data from ABCD with dorsal (superior) cutoff artifacts, which were excluded from the present study. Second, our synthetic data generation constrained cutting plane locations using two parameters: angles (-30 to 30 degrees) and vertical positions (-30 to -50 mm), based on real-life datasets. The impact of cutting plane locations on the performance of tractography parcellation merits further investigation. Future work is needed to systematically investigate cutting plane parameters needed for training to handle both inferior and superior FOV cutoffs, with additional testing in clinical datasets. Such efforts could provide deeper insights into the robustness of tractography parcellation methods and improve the practical utility of tractography parcellation in real-world applications. Third, we evaluated the performance of TractCloud-FOV on two incomplete FOV tractography datasets comprising children and young adults. To evaluate the generalizability of our proposed method, future research should identify suitable datasets for assessing the framework's performance across the lifespan, and investigate performance of the framework in the setting of pathology or neurodegeneration affecting brain connectivity [Essayed et al., 2017; Gatto et al., 2022; Gatto et al., 2024; van Gool et al., 2024; Yeh et al., 2021]. Finally, we used a single tractography method and one white matter tract atlas. Although comparisons of different tractography methods or atlases are beyond the scope of this study, we note that the proposed network is general and can be trained on other types of tractography and other atlases in the future.

In this work, we demonstrated the impact of inferior incomplete FOV artifacts on tractography parcellation and introduced a novel solution. We proposed TractCloud-FOV, a deep learning framework designed to robustly handle incomplete diffusion MRI fields of view during tractography parcellation. Central to our approach is the proposed FOV-Cut Augmentation (FOV-CA), which synthetically cuts tractograms to simulate varying degrees of real-world inferior FOV cutoffs. This strategy enables robust model training and evaluation under conditions mimicking real-world incomplete FOV cutoff challenges. Compared to SOTA methods, TractCloud-FOV demonstrates superior generalization across challenging, real-world datasets with incomplete fields of view, achieving more reliable tract identification, improved consistency relative to the training atlas, and enhanced computational efficiency. Our results demonstrate that TractCloud-FOV successfully addresses challenges posed by incomplete FOV in dMRI, achieving robust and accurate tractography parcellation.